\documentclass[twocolumn]{article}
\usepackage[letterpaper,top=2cm,bottom=2cm,left=2cm,right=2cm,marginparwidth=1.75cm]{geometry}
\usepackage{microtype}
\usepackage{graphicx}
\usepackage{graphics}
\usepackage{subfigure}
\usepackage{booktabs} 
\usepackage{caption}
\usepackage{hyperref}
\hypersetup{colorlinks,citecolor=blue, linkcolor=blue,urlcolor=blue}

\RequirePackage{algorithm,algorithmic}
\usepackage{amsmath}
\usepackage{amssymb}
\usepackage{mathtools}
\usepackage{amsthm}
\usepackage{float} 
\usepackage{bbm}
\usepackage{natbib}
\usepackage[parfill]{parskip}
\theoremstyle{plain}

\theoremstyle{definition}

\theoremstyle{remark}

\newtheorem{prop}{Proposition}
\newtheorem{cond}{A}
\usepackage{times}
\usepackage{changepage}
\usepackage{authblk}
\usepackage[textsize=tiny]{todonotes}

\title{
    \vspace{-1cm}
    \linethickness{1pt}\line(1,0){500}\\
    \vspace{0.5cm}
    \textbf{Transport Score Climbing: Variational Inference Using Forward KL and Adaptive Neural Transport}\\
    \linethickness{1pt}\line(1,0){500}
}

\author{\textbf{Liyi Zhang$^1$ \quad David M. Blei$^1$ \quad Christian A. Naesseth$^2$}}
\date{$^1\,$Columbia University \\
    $^2\,$University of Amsterdam \\
    zhang.liyi@columbia.edu \quad
    david.blei@columbia.edu \quad christian.a.naesseth@uva.nl
    }

\begin{document}

\maketitle

\begin{abstract}
\begin{adjustwidth}{0.6cm}{0.6cm}
Variational inference often minimizes the ``reverse'' Kullbeck-Leibler (KL) KL$(q||p)$ from the approximate distribution $q$ to the posterior $p$. Recent work studies the ``forward'' KL KL$(p||q)$, which unlike reverse KL does not lead to variational approximations that underestimate uncertainty. This paper introduces Transport Score Climbing (TSC), a method that optimizes KL$(p||q)$ by using Hamiltonian Monte Carlo (HMC) and a novel adaptive transport map. The transport map improves the trajectory of HMC by acting as a change of variable between the latent variable space and a warped space. TSC uses HMC samples to dynamically train the transport map while optimizing KL$(p||q)$. TSC leverages synergies, where better transport maps lead to better HMC sampling, which then leads to better transport maps. We demonstrate TSC on synthetic and real data. We find that TSC achieves competitive performance when training variational autoencoders on large-scale data.
\end{adjustwidth}
\end{abstract}

\section{Introduction}

A main goal in probabilistic modeling is to find the posterior distribution of latent variables given observed data. Probabilistic modeling allows using both structured knowledge and flexible parameterizations, including neural networks, but the posterior is often intractable. In this situation, we can use approximate inference to estimate the posterior distribution \citep{bishop:2006:PRML}.

Variational Inference (VI) is an optimization-based approximate inference method. It posits a family of distributions, and chooses a distribution $q$ in that family to approximate the posterior $p$ of a probabilistic model. It is a popular method for complex models because of its computational convenience, particularly when optimizing the ``reverse'', or ``exclusive'', Kullbeck-Leibler (KL) divergence KL$(q||p)$ through stochastic gradient descent (SGD) \citep{Jordan1999,Hoffman2013,blei2017variational}. 

However, reverse VI leads to approximations that may underestimate the uncertainty in $p$ \citep{minka2005divergence,Yao2018YesBD}. As an alternative, forward VI minimizes the ``forward'', or ``inclusive'', KL KL$(p||q)$. This approach better captures posterior uncertainty, but it is more computationally challenging \citep{Bornschein2015, gu2015neural, finke2019importanceweighted, Naesseth2020}. 

Another approach to approximate inference is Markov chain Monte Carlo (MCMC). MCMC methods sample from a Markov chain whose stationary distribution is the posterior, and produce good samples if run for long enough. However, in practice MCMC methods can be more computationally demanding than reverse VI in that they can take many iterations to converge.

To combine the advantages of both paradigms, \citet{Naesseth2020} introduce Markovian score climbing (MSC). MSC is a variational method for minimizing the forward KL($p||q$), which uses a Markov chain to approximate its intractable expectation over $p$. MSC uses an MCMC chain to approximate the expectation without asymptotic biases. However, this method uses basic MCMC kernels that can lead to slow exploration of the sampling space.

In this paper, we develop transport score climbing (TSC), a new algorithm that reliably and efficiently minimizes KL$(p||q)$. TSC uses
the MSC framework, but replaces the simple MCMC kernel with a Hamiltonian Monte Carlo (HMC) on a transformed, or warped, space \citep{Marzouk_2016, mangoubi2017rapid, hoffman2019neutralizing}. In particular, we adaptively transform the HMC sampling space, where the transformation is based on the current iteration of the variational approximation.

In more detail, TSC optimizes a normalizing flow \citep{rezende2015flow}, where the flow (or, equivalently, transport map) is trained from HMC samples from the warped space. Thus, TSC trains its transport map from scratch and leverages a synergy between the Markov chain and the variational approximation: an updated transport map improves the HMC trajectory, and the better HMC samples help train the transport map.

Finally, we show how TSC is amenable to SGD on large-scale IID data. To this end, we use TSC to improve training of deep generative
models with a variational autoencoder (VAE) \citep{Kingma2014AutoEncodingVB, Rezende2014StochasticBA}.

\paragraph{Contributions.}
1) We minimize KL$(p||q)$ with flow posteriors and an adaptive HMC kernel. The HMC kernel reuses the flow posterior to warp the underlying latent space for more efficient sampling. 2) Under the framework of VI with KL$(p||q)$, we show that the transport map of the warped space can be trained adaptively, instead of requiring a separate pre-training suggested by previous methods. 3) Empirical studies show that TSC more closely approximates the posterior distribution than both reverse VI and MSC. Furthermore, we use this methodology to develop a novel VAE algorithm competitive against four benchmarks, featuring continuously run HMC chains requiring no reinitializations, which are used by previous methods. 

\paragraph{Related Work.}

Forward VI is explored by several approaches. \citet{Bornschein2015,finke2019importanceweighted,jerfel2021variational} study VI with KL($p||q$) by using importance sampling (IS), and \citet{gu2015neural} uses sequential Monte Carlo (SMC). \citet{dieng2019reweighted} combines IS and VI in an expectation maximization algorithm. IS and SMC introduce a non-vanishing bias that leads to a solution which optimizes KL($p||q$) and the marginal likelihood only approximately \citep{naesseth2019elements,Naesseth2020}. Closest to the method proposed here is \citet{Naesseth2020,ou2020joint,gabrie2021adaptive}, which all use MCMC kernels to minimize KL($p||q$). \citet{ou2020joint,gabrie2021adaptive} can be considered to be instances of MSC \citep{Naesseth2020}. We build on MSC and propose to use the more robust HMC kernel together with a space transformation. The work of \citet{kim2022markov}, running parallel Markov chains for improved performance, can be combined with TSC for potential further gains.

\citet{mangoubi2017rapid} show that MCMC algorithms are more efficient on simpler spaces, such as on strongly log-concave targets. \citet{Marzouk_2016,hoffman2019neutralizing} use transformations to create warped spaces that are easy to sample from. The transformation is defined by functions called ``transport maps'' that are pre-trained by reverse VI. The proposed algorithm differs in the optimization objective and by learning the transport map together with model parameters end-to-end.

Using MCMC to learn model parameters based on the maximum marginal likelihood is studied in many papers, e.g., \citet{Gu7270,KuhnL:2004,andrieu2006,AndrieuV:2014}. In contrast TSC proposes a new method for the same objective, by adapting the MCMC kernel using VI.

\citet{Kingma2014AutoEncodingVB, Rezende2014StochasticBA} introduce variational autoencoders (VAE) where both the generative model and the approximate posterior are parameterized by neural networks. They optimize a lower bound to the marginal log-likelihood, called the evidence lower bound (ELBO), with the reparameterization trick.  \citet{salimans2015markov, caterini2018hamiltonian} incorporate MCMC or HMC steps to train a modified ELBO with lower variance. \citet{pmlr-v70-hoffman17a, hoffman2019neutralizing, ruiz2021unbiased} instead formulate the optimization as maximum likelihood while utilizing MCMC methods. The work proposed here also targets maximum likelihood, but we neither augment the latent variable space \citep{ruiz2021unbiased} nor reinitialize the Markov kernel from the posterior at each epoch \citep{pmlr-v70-hoffman17a, hoffman2019neutralizing}. Instead, we continuously run the Markov kernel on the warped latent variable space.

\section{Background}

Let $p({\bf x,z})$ be a probabilistic model, with $\textbf{z}$ as latent variables and $\textbf{x}$ as observed data. A main goal of Bayesian inference is to calculate or approximate the posterior distribution of latent variables given data, $p({\bf z|x})$.

VI approximates the posterior by positing a family of distributions $\mathcal{Q}$, where each distribution takes the form $q(\textbf{z};\lambda)$ with variational parameters $\lambda$. The most common approach, reverse VI, minimizes the reverse KL using gradient-based methods: $\min_{\lambda} \text{KL}(q(\textbf{z};\lambda)||p(\textbf{z}|\textbf{x}))$. The main strength of reverse \textsc{VI} is computational convenience. 

\subsection{Variational Inference with Forward KL}

Reverse \textsc{VI} often underestimates the uncertainty in $p$. An alternative approach, which is the focus of this work, is to minimize the forward KL: $\min_{\lambda}\text{KL}(p(\textbf{z}|\textbf{x})||q(\textbf{z};\lambda))$. While more challenging to work with, this objective does not lead to approximations that underestimate uncertainty \citep{Naesseth2020}. Moreover, if $\mathcal{Q}$ is a subset of the exponential family distributions, the moments of the optimal $q$ matches the moments of the posterior $p$ exactly. 

The forward KL divergence from $p$ to $q$ is
\begin{align} \label{eq:1}
    \text{KL}(p(\textbf{z}|\textbf{x})||q(\textbf{z};\lambda)) := \mathbb{E}_{p(\textbf{z}|\textbf{x})}\left[
    \log \frac{p(\textbf{z}|\textbf{x})}{ q(\textbf{z};\lambda)}\right]. 
\end{align}
To minimize eq. (\ref{eq:1}), the gradient w.r.t. the variational parameters is,
\begin{align} \label{eq:2}
    \mathbb{E}_{p({\bf z|x})} [ -\nabla_{\lambda} \log q(\textbf{z};\lambda) ].
\end{align}
Approximating the expectation over the unknown posterior $p({\bf z|x})$ is a major challenge. \citet{Bornschein2015,gu2015neural} approximate the expectation in eq. (\ref{eq:2}) through importance sampling and sequential Monte Carlo, but these methods gives estimates of the gradient with systematic bias. 

In this work we leverage Markovian score climbing (\textsc{MSC}) \citep{Naesseth2020}, which uses samples $\textbf{z}$ from an \textsc{MCMC} kernel with the posterior $p({\bf z|x})$ as its stationary distribution. The resulting SGD method leads to an algorithm that provably minimizes $\text{KL}(p||q)$ \citep{Naesseth2020}. 

\paragraph{Normalizing Flow.}
In this work we focus on the variational family of normalizing flows. Normalizing flows transform variables with simple distributions and build expressive approximate posteriors \citep{rezende2015flow,tabak2013family}, and are tightly linked with warped space HMC. Given a $d$-dimensional latent variable \textbf{z}, the transformation uses an invertible, smooth, trainable function $T_{\lambda}: \mathbb{R}^d \mapsto \mathbb{R}^d$ and introduces a random variable $\boldsymbol{\epsilon}$ with a simple distribution $q_0(\boldsymbol{\epsilon})$, oftentimes an isotropic Gaussian. Using the change-of-variable identity, the probability density function $q$ of $T_{\lambda}(\boldsymbol{\epsilon})$ is,
$$q(T_{\lambda}(\boldsymbol{\epsilon})) = q_0(\boldsymbol{\epsilon}) \Big\lvert \text{det} \frac{dT_{\lambda}}{d\boldsymbol{\epsilon}} \Big\rvert^{-1},$$
where $\frac{dT_{\lambda}}{d\boldsymbol{\epsilon}}$ is the Jacobian matrix.

\subsection{Neural Transport HMC}

The HMC kernel used in the algorithm proposed below is closely related to Neural Transport HMC (NeutraHMC), proposed by \citet{hoffman2019neutralizing}. NeutraHMC simplifies the geometry of the sampling space through neural network-parameterized transport maps. Compared to HMC, it explores the target distribution more efficiently. We briefly explain HMC and NeutraHMC.

\paragraph{Hamiltonian Monte Carlo.}
HMC is an MCMC algorithm that produces larger moves in \textbf{z} by introducing ``momentum'' variables \textbf{m} of the same dimension \citep{duane-mcem, neal-hmc}. It constructs a joint proposal on the augmented space $(\textbf{z}, \textbf{m})$ to target $p(\textbf{z}|\textbf{x}) p(\textbf{m})$. A common choice for the distribution $p(\textbf{m})$ is $N(0, I)$.

In a given iteration, a proposal involves $L$ ``leapfrog steps'' of step-size $s$ each defined by
\begin{align*}
    \textbf{m} &= \textbf{m} + \frac{1}{2}s\frac{d\log p(\textbf{x},\textbf{z})}{d\textbf{z}}, \\
    \textbf{z} &= \textbf{z} + sM^{-1}\textbf{m}, \\
    \textbf{m} &= \textbf{m} + \frac{1}{2}s\frac{d\log p(\textbf{x},\textbf{z})}{d\textbf{z}},
\end{align*}
starting from $(\textbf{z},\textbf{m})$, $\textbf{m} \sim N(0, I)$. The final leapfrog step gives the proposed state $(\widetilde{\textbf{z}}, \widetilde{\textbf{m}})$. The new state is accepted with probability $\displaystyle{\min \{1, \frac{p(\textbf{x},\widetilde{\textbf{z}})p( \widetilde{\textbf{m}})}{p(\textbf{x},\textbf{z}) p(\textbf{m})} \}}$ \citep{neal-hmc,robert2013monte}. 

\paragraph{HMC on Warped Space.}
\citet{Marzouk_2016, mangoubi2017rapid, hoffman2019neutralizing} propose running MCMC methods on a simpler geometry through transforming, or warping, the sampling space with a transport map. A transport map is defined as a parameterized function $T_{\lambda}(\cdot)$. The warped space is defined by the change in variable  $\boldsymbol{z_0} = T_{\lambda}^{-1}(\textbf{z})$ for $\textbf{z}\sim p(\textbf{z}|\textbf{x})$. If $T_{\lambda}$ is chosen well, $\boldsymbol{z_0}$ will be simpler than \textbf{z} to sample. The target distribution in the MCMC algorithm is defined as the distribution of $\boldsymbol{z_0}$. Each $\boldsymbol{z_0}^{(k)}$ generated by MCMC at the $k$-th iteration is passed to the transport map with $\textbf{z}^{(k)} = T_{\lambda}(\boldsymbol{z_0}^{(k)})$. $(\textbf{z}^{(1)},\textbf{z}^{(2)},...)$ then have the true target distribution as its stationary distribution, but with faster mixing than MCMC on the original space. \citet{hoffman2019neutralizing} introduces NeutraHMC that uses HMC instead of general MCMC. NeutraHMC utilizes both affine and neural network transport maps that are pretrained using VI based on KL$(q||p)$.

\section{Transport Score Climbing}

\begin{figure}[t]
    \centering
    \includegraphics[width=0.35\textwidth]{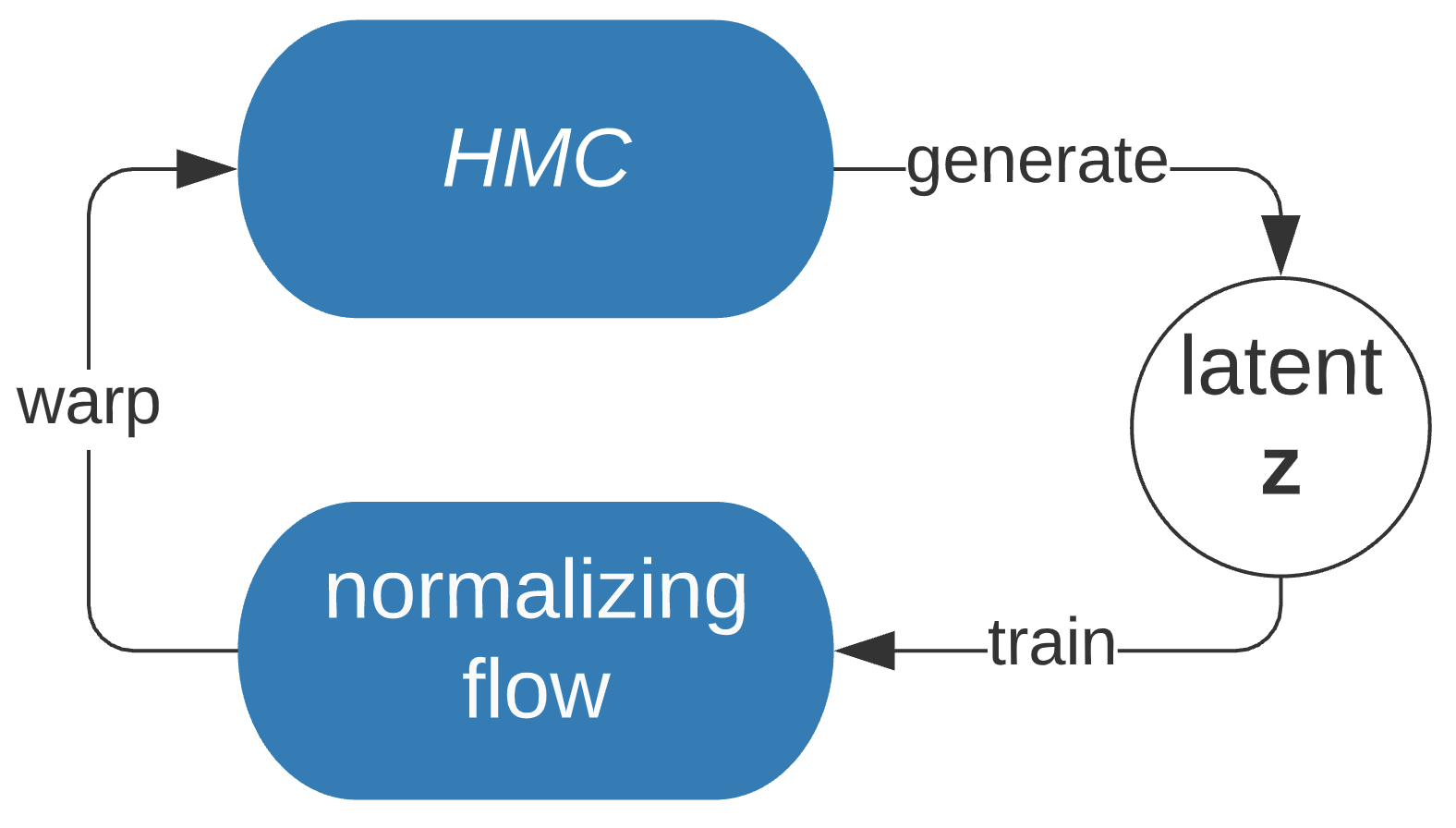}
    \caption{Outline of the TSC algorithm. HMC generates samples of latent variable \textbf{z} that are used to train the normalizing flow transformation. The refined transformation further improves the geometry of HMC, which goes on to generate the next sample.}
    \label{fig:tsc-flowchart}
\end{figure}

We now develop Transport Score Climbing (\textsc{TSC}), a method for \textsc{VI} with forward KL$(p||q)$. TSC uses HMC on warped space to estimate the intractable expectation in the gradient (eq.(\ref{eq:2})). A transport map is defined to act as both flow transformation in the variational posterior $q$ and mapping between HMC sampling spaces. As $q$ is updated, the mapping is updated simultaneously which further refines the HMC sampling space. Figure \ref{fig:tsc-flowchart} shows the synergy between HMC sampling and variational approximation. Under conditions (Section \ref{sec:convergence}), TSC converges to a local optimum of KL$(p||q)$. 

\subsection{Types of Transport Maps}

Let $\boldsymbol{\epsilon} \sim N(0, I_d)$, let $T_{\lambda}(\cdot)$ be a function $\mathbb{R}^d \mapsto \mathbb{R}^d$, or transport map, with trainable parameter $\lambda$, and define the variational distribution $q(\textbf{z};\lambda)$ such that $\textbf{z} = T_{\lambda}(\boldsymbol{\epsilon}) \sim q(\textbf{z};\lambda)$. The variational distribution and transport map share trainable parameters. We consider three concrete examples.

\paragraph{Affine Transformation.} 
Consider an affine transformation $T_{\lambda}(\boldsymbol{\epsilon}) = \boldsymbol{\mu} + \boldsymbol{\sigma} \odot\boldsymbol{\epsilon}$, where $\odot$ denotes elementwise multiplication. The variational distribution is $q(\textbf{z};\lambda) = N(\textbf{z};\boldsymbol{\mu}, \boldsymbol{\sigma}^2 I)$. In the empirical studies, Section~\ref{sec:expts}, we find that the affine transport map is simple and effective.

\paragraph{IAF Transformation.} 
A popular flow transformation is the inverse autoregressive flow (IAF) \citep{kingma-iaf}. $T_{\lambda}$ is chosen with the autoregressive property, that is, along each dimension $i$ of $\boldsymbol{\epsilon}$,
$$T_i(\boldsymbol{\epsilon}) = \boldsymbol{\epsilon}_i \sigma_i(\boldsymbol{\epsilon}_{1:i-1};\phi) + \mu_i(\boldsymbol{\epsilon}_{1:i-1};\phi).$$
Here $\mu$ and $\sigma$ are neural networks that act as shift and scale functions. IAF is flexible because of neural networks; its determinant is cheap to compute because of the autoregressive property that leads to a lower triangular Jacobian matrix. However, the inverse IAF $T^{-1}_{\lambda}(\textbf{z})$, required to evaluate the density $q$, is costly to compute. Thus, we only use IAF in studies where latent variables are low-dimensional.

\paragraph{RealNVP Transformation.} RealNVP is an alternative to IAF with slightly less flexibility for the same number of parameters but fast invertibility \citep{dinh-realnvp}. 
RealNVP uses an affine coupling layer to transform the input $\boldsymbol{\epsilon}$. In practice, we use a checkerboard binary mask $b$ to implement the transformation, as detailed in \citet{dinh-realnvp},
$$T(\boldsymbol{\epsilon}) = b \odot \boldsymbol{\epsilon} + (1-b) \odot \Big( \boldsymbol{\epsilon} \odot \text{exp}\big( \sigma(b \odot \boldsymbol{\epsilon}) \big) + \mu(b \odot \boldsymbol{\epsilon}) \Big).$$
where $\mu$ and $\sigma$ are also neural networks. The idea is that the part of $\boldsymbol{\epsilon}$ that is transformed by neural networks depend only on the other part of $\boldsymbol{\epsilon}$ that goes through the identity function. This construction allows for fast inversion. 

Both IAF and RealNVP flow transformations can be stacked to form more expressive approximations. Let $T_{\lambda}^{(l)}$ denote one IAF or RealNVP transformation. We stack L transformations, and define the transport map as ${\displaystyle T_{\lambda}(\boldsymbol{\epsilon}) = T_{\lambda_L}^{(L)} \circ ... \circ T_{\lambda_1}^{(1)}(\boldsymbol{\epsilon})}$. The variational distribution $q(\textbf{z};\lambda)$ is a flow-based posterior with ${\displaystyle q(\textbf{z};\lambda) = N(\boldsymbol{\epsilon}; 0, I) \Big\lvert \text{det} \frac{d T_{\lambda}(\boldsymbol{\epsilon})}{d \boldsymbol{\epsilon}}\Big\rvert^{-1}}$.

\begin{algorithm}[t]
   \caption{\textsc{TSC}}
   \label{alg:hsc}
   \textbf{Input:} Probabilistic model $p(\textbf{z},\textbf{x};\theta)$; transformation $T_{\lambda}: \mathbb{R}^d \mapsto \mathbb{R}^d$; \textsc{HMC} kernel $H[\boldsymbol{z_0}^{(k+1)}|\boldsymbol{z_0}^{(k)};\lambda,\theta]$ with target distribution $p(\boldsymbol{z_0}|\textbf{x};\theta, \lambda)$ and initial state $\boldsymbol{z_0}^{(0)}$; variational distribution $q(\textbf{z};\lambda)$; step-sizes $\alpha_1, \alpha_2$. $\lambda, \theta$ randomly initialized. \\
   \textbf{Output:} $\lambda, \theta$.
   \begin{algorithmic}
   \FOR{$k \in \{0,1,2,...\}$}
   \STATE $\boldsymbol{z_0}^{(k+1)} \sim H[\boldsymbol{z_0}^{(k+1)}|\boldsymbol{z_0}^{(k)};\lambda,\theta]$.
   \STATE $\textbf{z} = T_{\lambda}(\boldsymbol{z_0}^{(k+1)})$.
   \STATE $\textbf{z} = \text{stop-gradient}(\textbf{z})$.
   \STATE $\lambda = \lambda - \alpha_1 \nabla_{\lambda} [-\log q(\textbf{z};\lambda)]$.
   \STATE $\theta = \theta - \alpha_2 \nabla_{\theta}[-\log p(\textbf{x}|\textbf{z};\theta) - \log p(\textbf{z})]$.
   \STATE $\boldsymbol{z_0}^{(k+1)} = T_{\lambda}^{-1}(\textbf{z})$.
   \ENDFOR
   \end{algorithmic}
   \label{alg:tsc}
\end{algorithm}

\subsection{VI with HMC on Warped Space}

In order to sample the latent variables $\textbf{z}$, we define the target of the HMC kernel $H(\cdot | \boldsymbol{z_0})$ as the distribution of $\boldsymbol{z_0} = T_{\lambda}^{-1}(\textbf{z})$, $\textbf{z} \sim p(\textbf{z}|\textbf{x})$,
\begin{align}
    p(\boldsymbol{z_0} | \textbf{x}; \lambda) &\propto
    p(\textbf{x}, T_{\lambda}(\boldsymbol{z_0})) |\text{det}J_{T_{\lambda}}(\boldsymbol{z_0})|, \label{eq:3}
\end{align}
where $J_f(x)$ is the Jacobian matrix of function $f$ evaluated at $x$.
This means that we are sampling on the warped space defined by $T_{\lambda}(\cdot)$ rather than the original space of latent variables. After $\boldsymbol{z_0}$ is sampled, we pass it to the transport map with $\textbf{z} = T_{\lambda}(\boldsymbol{z_0})$ to acquire the latent variable sample. 
As in MSC \citep{Naesseth2020}, we do not re-initialize the Markov chain at each iteration, but use the previous sample $\textbf{z}^{(k)}$ to both estimate the gradient and serve as the current state of the HMC kernel $H$ to sample $\textbf{z}^{(k+1)}$.

A crucial element is that the transport map  $T_{\lambda}$ is trained jointly as we update the KL$(p||q)$ objective in eq.(\ref{eq:2}). This is because the map is also the flow transformation part of the variational distribution $q$. Specifically, HMC at iteration $k$ uses variational parameters of the previous iteration, $\lambda^{(k-1)}$, in its target distribution (eq.(\ref{eq:3})) at iteration $k$. By construction, if $q$ is close to the true posterior, target $p(\boldsymbol{z_0} | \textbf{x}; \lambda)$ will be close to the isotropic Gaussian. Therefore, TSC keeps refining the geometry of the HMC sampling space throughout the training process.

\subsubsection{Model Parameters}

The probabilistic model $p({\bf z, x}; \theta)$ can also contain unknown parameters $\theta$. The corresponding warped space posterior is
\begin{align}
    p(\boldsymbol{z_0} | \textbf{x}; \lambda,\theta) &\propto
    p(\textbf{x}, T_{\lambda}(\boldsymbol{z_0});\theta) |\text{det}J_{T_{\lambda}}(\boldsymbol{z_0})|.
    \label{eq:model}
\end{align}
Taking samples from the true posterior $p({\bf z|x; \theta})$ allows one to learn $\theta$ using maximum likelihood, optimizing the marginal likelihood $p(\textbf{x}; \theta)$. This fact follows from the Fisher identity, which writes the gradient of the marginal likelihood as an expectation over the posterior,
\begin{align} 
    \nabla_{\theta}\log p(\textbf{x}; \theta) &= \nabla_{\theta}\log \int p({\bf z, x}; \theta)d\textbf{z} \nonumber\\
    &= \mathbb{E}_{p({\bf z|x; \theta})}[\nabla_{\theta}\log p({\bf z, x};\theta)]. 
    \label{eq:4}
\end{align}
The expectation above is estimated by the same HMC sample $\textbf{z}^{(k)}$ that is used to update variational parameters $\lambda$. Additionally, the HMC kernel at iteration $k$ uses model parameters of the previous iteration, $\theta^{(k-1)}$, in its target distribution (eq.(\ref{eq:model})) at iteration $k$. The corresponding algorithm can be shown to maximize the true marginal likelihood exactly, see e.g. \citep{Gu7270} and Section~\ref{sec:convergence}. 

Algorithm \ref{alg:tsc} summarizes TSC for learning $\lambda$ and $\theta$.

\subsection{Amortized Inference}
When the dataset $\textbf{x} = (\textbf{x}_1,...,\textbf{x}_n)$ is i.i.d. with empirical distribution $\widehat{p}(\textbf{x})$ each $\textbf{x}_i$ has its own latent variable $\textbf{z}$. Amortized inference then uses the approximate posterior $q(\textbf{z}|\textbf{x};\lambda)$ instead of a separate $q(\textbf{z};\lambda_i)$ for each $\textbf{x}_i$. In amortized inference, variational parameters $\lambda$ are shared across data-points $\textbf{x}_i$. It is known as a VAE when both the likelihood $p(\textbf{x}|\textbf{z};\theta)$ and the approximate posterior $q$ are parameterized by neural networks. 

TSC conducts maximum likelihood and VI with KL$(p||q)$ on VAE and is amenable to SGD with mini-batches. Following derivations from \citet{Naesseth2020}, the gradient with respect to $\lambda$ is 
\begin{align}
    &\mathbb{E}_{\widehat{p}(\textbf{x})}\big[ \nabla_{\lambda}\text{KL}(p(\textbf{z}|\textbf{x};\theta) || q(\textbf{z}|\textbf{x};\lambda)) \big] \nonumber\\ 
    &\approx \frac{1}{M} \sum_{i=1}^M \mathbb{E}_{p(\textbf{z}|\textbf{x}_i)}[-\nabla_{\lambda} \log q(\textbf{z}|\textbf{x}_i;\lambda)], \label{eq:5}
\end{align}
where $M$ is the mini-batch size. For model learning, we similarly estimate the gradient using eq. (\ref{eq:4}),
\begin{align}
    &\mathbb{E}_{\widehat{p}(\textbf{x})}\big[\nabla_{\theta}\log p(\textbf{x}; \theta)\big] \nonumber \\
    &= \mathbb{E}_{\widehat{p}(\textbf{x})}\big[\mathbb{E}_{p(\textbf{z}|\textbf{x};\theta)}[\nabla_{\theta}[\log p(\textbf{x}|\textbf{z};\theta) + \log p(\textbf{z})]]\big] \nonumber\\
    &\approx \frac{1}{M} \sum_{i=1}^M \mathbb{E}_{p(\textbf{z}|\textbf{x}_i;\theta)}[\nabla_{\theta}[\log p(\textbf{x}_i|\textbf{z};\theta) + \log p(\textbf{z})]]. \label{eq:6}
\end{align}
The expectations are approximated using HMC samples, as in Algorithm \ref{alg:tsc}. Similarly with the non-amortized case, we do not re-initialize the Markov chain at each iteration, but approximate the expectation by running one step of the Markov chain on the previous sample $\textbf{z}^{(k-1)}$.

\newdimen\figrasterwd
\figrasterwd\textwidth

\begin{figure*}[h]
\parbox{\figrasterwd}{
    \parbox{0.68\figrasterwd}{
        \subfigure[First dimension.]{\includegraphics[width=5.5cm,height=5cm]{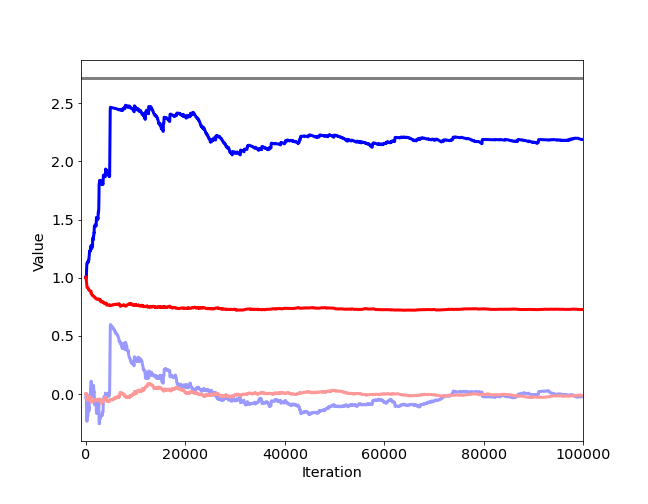}}
        \subfigure[Second dimension.]{\includegraphics[width=5.5cm,height=5cm]{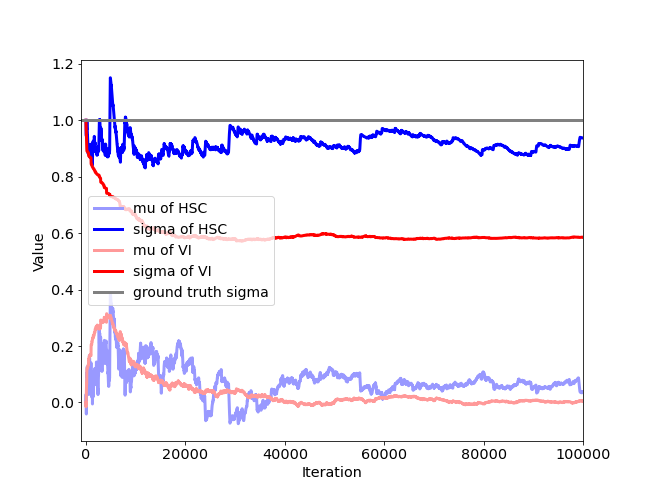}}
        \caption{Diagonal Gaussian variational parameters in TSC and ELBO VI across iterations, approximating the funnel distribution. Each subplot shows parameter values of one dimension of the distributions. Ground truth $\sigma$ is also drawn, but ground truth $\mu=0$ is not drawn here. \emph{TSC, while slightly more volatile than ELBO maximization, converges to parameter values closer to the ground truth.}}
        \label{fig:funnel-gaussian-conv}
    }
    \parbox{0.01\figrasterwd}{\(\)}
    \parbox{0.28\figrasterwd}{
    \hspace{0.8cm}
        \subfigure{
        \includegraphics[width=3cm]{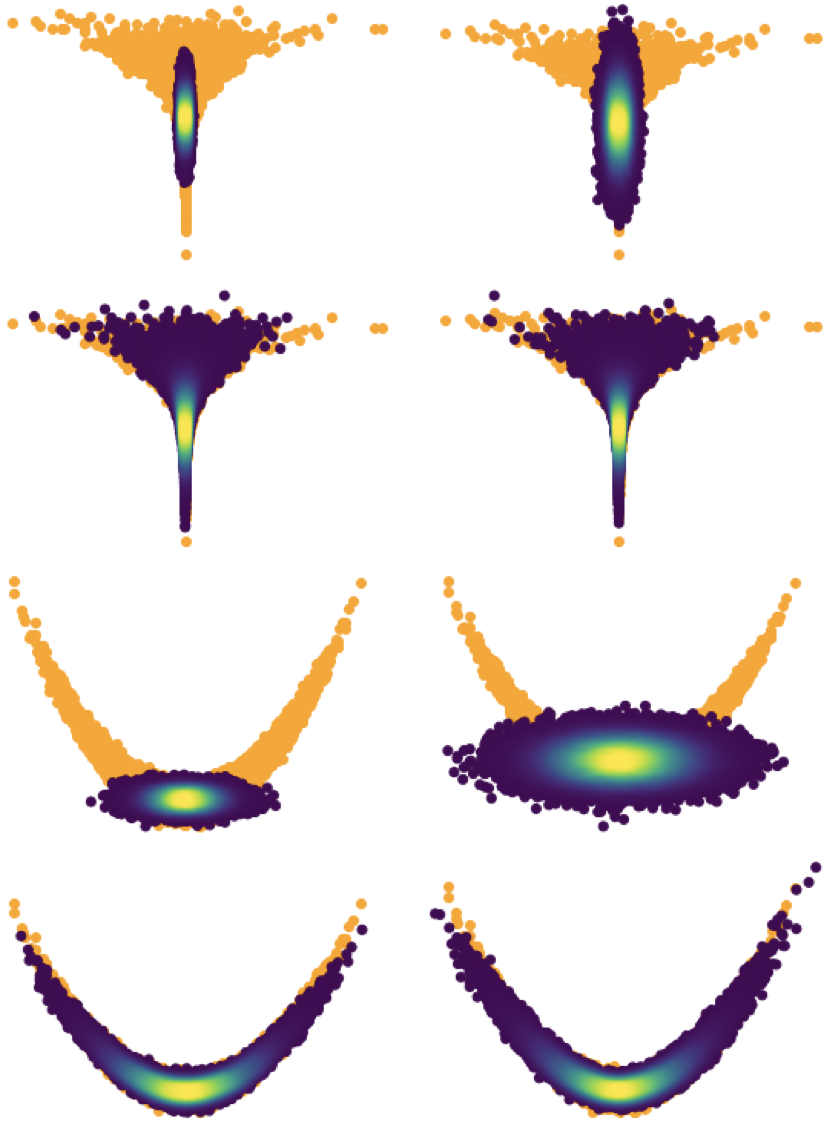}}
        \caption{Synthetic targets (orange) with fitted posteriors laid on top. Top two: funnel; bottom two: banana. Rows 1 \& 3: Gaussian family; rows 2 \& 4: IAF family. Left: VI; right: TSC. 
        \emph{TSC more accurately approximates the posterior distribution.}
        } 
        \label{fig:synthetic-dist}
    }
}
\end{figure*}

\section{Empirical Evaluation}\label{sec:expts}

All implementations are made in TensorFlow and TensorFlow Probability \citep{tensorflow2015-whitepaper, DBLP:journals/corr/abs-1711-10604}.\footnote{Code is available at \url{https://github.com/zhang-liyi/tsc}.} On two synthetic datasets, TSC converges to near-optimal values. On survey data, TSC is more efficient than MSC and gives more reliable approximations on this task. For VAE, TSC achieves higher log-marginal likelihood on static MNIST, dynamic MNIST, and CIFAR10 than VAEs learned using four other baselines. 

\subsection{Synthetic Data}

\paragraph{Neal's Funnel Distribution.}

We first study the funnel distribution described by \citet{neal-2003}, a distribution known to be hard to sample from by \textsc{HMC}. Let random variable $\boldsymbol{z}$ have probability density function
$$ p(\boldsymbol{z}) = \mathcal{N}(z_1 | 0, 1)  \mathcal{N}(z_2 | 0, e^{z_1/2}).$$
Then, $\boldsymbol{z}$ follows the Funnel distribution.

\paragraph{Banana Distribution.}

Following \citet{haario1999}, we twist the Gaussian distribution to create a banana-shaped distribution. Let ${\textstyle (v_1, v_2) \sim \mathcal{N}(0, \begin{psmallmatrix} 100 & 0 \\ 0 & 1 \end{psmallmatrix})}$, we transform $(v_1, v_2)$ with,
\begin{align*}
    z_1 &= v_1, \\
    z_2 &= v_2 + b\cdot v_1^2 - 100b,
\end{align*}
where $b$ is a factor set to 0.02. Then, $(z_1, z_2)$ follows the Banana distribution. 

Both distributions are visualized in Figure \ref{fig:synthetic-dist}. We use the Adam optimizer \citep{Kingma2015AdamAM} with inverse time decay, decay rate $3\cdot10^{-4}$, and initial learning rate $3\cdot10^{-3}$. The HMC sampler consists of 1 chain, with step size $s$ tuned in $[0.03,1)$ to target 67\% acceptance rate, and number of leapfrog steps $L$ set to $\lceil \frac{1}{s} \rceil$.

\begin{table}[h]
\caption{Uncertainty estimation by the IAF family on synthetic data. The table gives standard deviation (std) across 100 groups, each group containing $10^7$ i.i.d. samples from fitted posteriors. In parentheses are standard errors across the 100 groups. \emph{All methods give reasonable approximations using an expressive distribution, but TSC more closely recovers the true target distribution.}}
\centering
\subtable[Funnel distribution.]{
\centering
\resizebox{0.8\columnwidth}{!}{
 \begin{tabular}{ccc} 
 \toprule
 Method & Std on Dim 1 & Std on Dim 2 \\
 \midrule
 Ground truth & 2.718 & 1 \\
 \midrule
 ELBO VI & 2.286 (0.002) & 0.989 (0) \\
 MSC & 2.151 (0.001) & 0.961 (0) \\
 \textbf{TSC} & \textbf{2.426 (0.002)} & \textbf{0.991 (0)} \\
 \bottomrule
\end{tabular}}}
\subtable[Banana distribution.]{
\centering
\resizebox{0.8\columnwidth}{!}{
 \begin{tabular}{ccc} 
 \toprule
 Method & Std on Dim 1 & Std on Dim 2 \\
 \midrule
 Ground truth & 10 & 3 \\
 \midrule
 ELBO VI & 9.511 (0.002) & 2.675 (0.001) \\
 MSC & 9.661 (0.002) & 2.562 (0.001) \\
 \textbf{TSC} & \textbf{9.949 (0.002)} & \textbf{2.883 (0.001)} \\
 \bottomrule
\end{tabular}}}
\label{tab:iaf}
\end{table}

\paragraph{Results.}

For the first variational family, we consider a diagonal Gaussian, $q(\boldsymbol{\theta}) = N(\boldsymbol{\theta} | \boldsymbol{\mu}, \boldsymbol{\sigma}^2 I)$. The optimal variational parameter for TSC is the true mean and standard deviation of $\boldsymbol{\theta}$.  

Figure \ref{fig:funnel-gaussian-conv} show variational parameters of the two dimensions by iteration. While both VI and TSC converge to near-optimal values for $\boldsymbol{\mu}$, VI significantly underestimates uncertainty by converging to low values of $\boldsymbol{\sigma}$. This problem is ameliorated by TSC, which gives $\boldsymbol{\sigma}$ estimates much closer to the ground truth. 


As a second variational approximation, we use an IAF with 2 hidden layers. With an expressive posterior, each method gives reasonable approximations (Figure \ref{fig:synthetic-dist}). Table \ref{tab:iaf} quantitatively compares these methods on synthetic data by giving standard deviations of large numbers of samples from the fitted IAF posteriors, and estimates from TSC are closest to the ground truth. 

However, TSC still gives approximations that are often a little narrower than the true target distribution. One reason is the difficulty of the HMC chain to sample from certain areas in the target distribution. While an expressive flow further simplifies the geometry for HMC, it still does not guarantee perfect approximation in finite time. 

\begin{figure*}
    \centering
    \includegraphics[width=15cm,height=5.5cm]{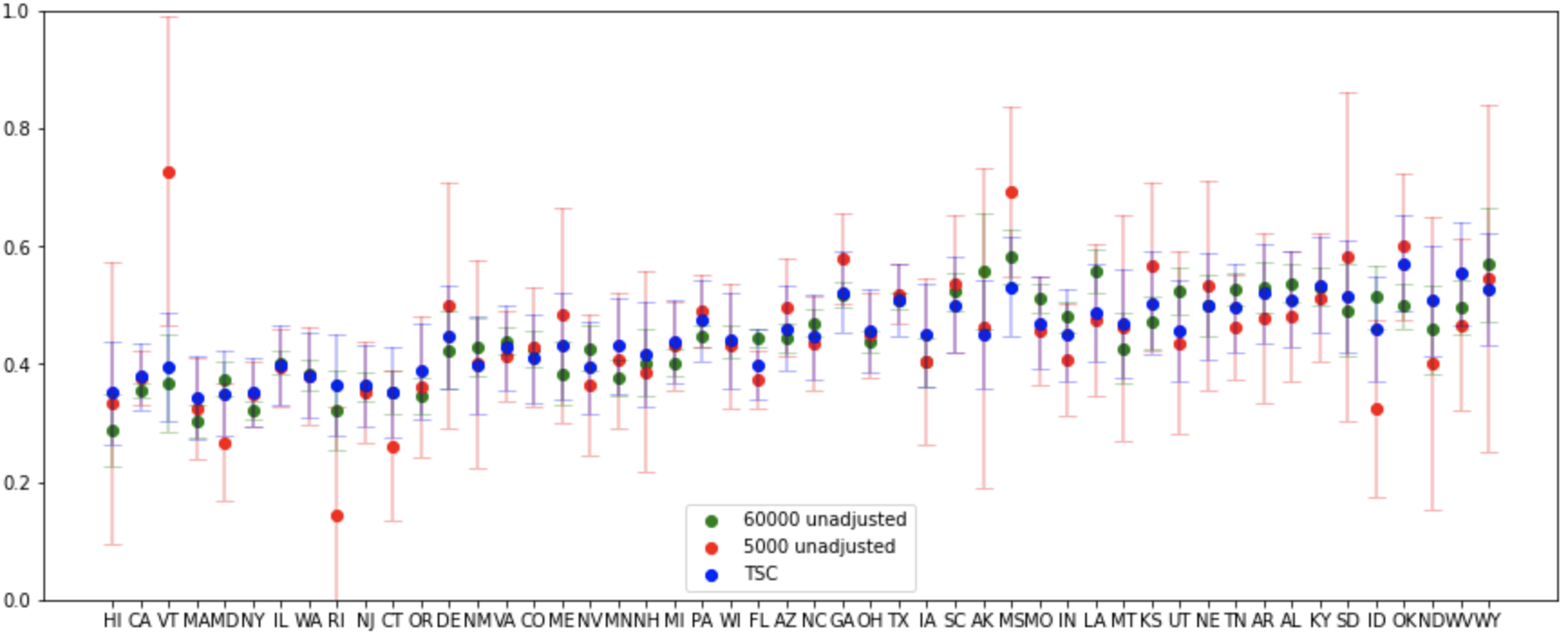}
    \caption{Estimates by states, where states are ordered by Republican vote in the 2016 election. We show TSC, 5000 sample unadjusted estimates, and 60000 sample unadjusted estimates. The unadjusted estimates are found by caluclating the mean and standard error of Bernoulli random variables. 95\% confidence intervals are plotted. The closer to green the better. \emph{TSC is robust to noise in the data sample and improves over the 5000 sample unadjusted estimate.}}
    \label{fig:survey-by-state}
\end{figure*}

\subsection{Survey Data}

We use the multilevel model from \citet{lopez-martin_phillips_gelman_2021} and apply it to a survey dataset provided by the same authors. Details of the model are given in Supplement \ref{supp:survey-model}. The dataset originally comes from the 2018 Cooperative Congressional Election Study \citep{DVN/ZSBZ7K_2019}. The survey response is a binary variable representing individual answers to the question: allow employers to decline coverage of abortions in insurance plans (Support / Oppose), and to each response is attached a set of features concerning the individual. The dataset consists of 60,000 data-points, but as suggested by the study, inference methods are trained on a randomly selected 5,000 subset. Reliable estimations are demonstrated by the ability to generalize from the 5,000 subset to the full 60,000 set, and closeness to gold-standard MRP MCMC results \citep{lopez-martin_phillips_gelman_2021}.

We implement MSC and TSC with diagonal Gaussian approximations, and train them using the Adam optimizer with inverse time decay, decay rate $10^{-3}$, and initial learning rates 0.01. The HMC sampler consists of 1 chain, with step size $s$ tuned in $[0.03,1]$ and number of leapfrog steps $L$ set to $\lceil \frac{1}{s} \rceil$.

Figure \ref{fig:survey-by-state} shows estimates by individuals' U.S. state, with states ordered by Republican vote share. The large-sample (60,000) estimates show an upward trend, which is intuitive. The estimates for TSC comes from 10,000 posterior samples. Figure \ref{fig:survey-by-state} shows that TSC gives reasonable approximations since it generalizes from the 5,000 data points and gives estimates that are robust against noise in the data sample and are close to 60,000 sample estimates. 

\begin{figure}
    \centering
    \includegraphics[width=0.38\textwidth]{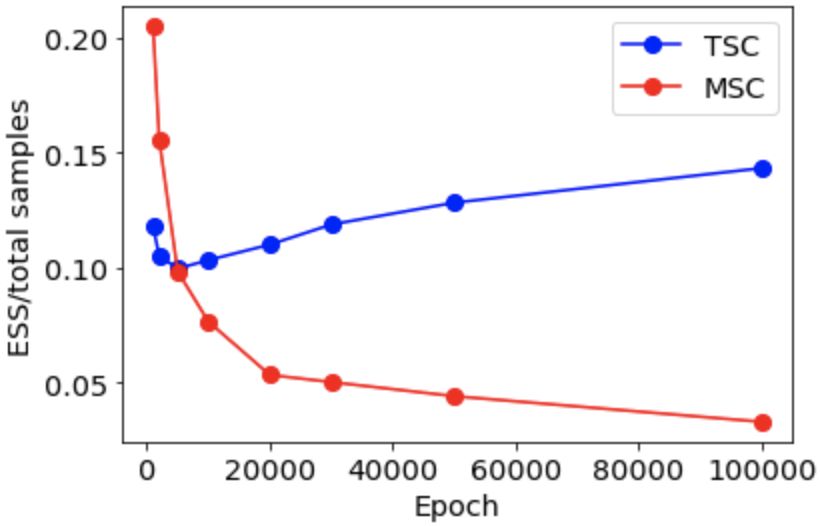}
    \caption{Cumulative ESS of TSC and MSC over number of samples across epochs. Higher is better. \emph{As the transport map is learnt, TSC achieves higher ESS.}}
    \label{fig:survey-ess}
\end{figure}

Asymptotic sample behavior measured through effective sample size (ESS) shows that the warped HMC chain underlying TSC outperforms the vanilla HMC chain used by MSC (Figure \ref{fig:survey-ess}). It also suggests that dynamic training of the transport map actually hurts HMC efficiency when the variational approximation is still poor, but it quickly catches up when the approximation is better trained.

ELBO VI and MSC also provide reasonable approximations, but TSC is closer to MRP MCMC. We quantitatively compare TSC, ELBO VI, and the gold-standard MRP MCMC estimates \citep{lopez-martin_phillips_gelman_2021}. Table \ref{tab:survey} shows that TSC results are closer to MRP MCMC results than VI is to MRP MCMC.

\begin{table}
\centering
\caption{The sum of squared difference from MRP MCMC estimates of mean and std of the response variable, one row for each method. \emph{Lower is better.}}
\resizebox{0.8\columnwidth}{!}{
 \begin{tabular}{cccc} 
 \toprule
 Method & Mean Difference & Std Difference\\ 
 \midrule
 ELBO VI & $1.18 \cdot 10^{-3}$ & $1.95 \cdot 10^{-3}$\\
 MSC & $2.44 \cdot 10^{-2}$ & $1.53 \cdot 10^{-2}$\\
 \textbf{TSC} & $\boldsymbol{5.86 \cdot 10^{-4}}$ & $\boldsymbol{1.02 \cdot 10^{-3}}$ \\
 \bottomrule
\end{tabular}}
\label{tab:survey}
\end{table}

\subsection{Variational Autoencoders}

Finally, we study TSC with amortized inference on statically binarized MNIST, dynamically binarized MNIST, and CIFAR10. With KL$(p||q)$ and dynamic updates of transport maps, a continuously run TSC is able to achieve higher log-marginal likelihood than several benchmarks.

\paragraph{Implementation Details.}
For benchmark methods, we use ELBO VI \citep{Kingma2014AutoEncodingVB, Rezende2014StochasticBA}, importance-weighted (IW) autoencoder \citep{Burda2016ImportanceWA}, MSC with the conditional importance sampler (CIS-MSC) \citep{Naesseth2020}, and NeutraHMC that follows the training procedure detailed in \citet{pmlr-v70-hoffman17a, hoffman2019neutralizing}. We use the Adam optimizer with learning rates 0.001 and mini-batch size 256. Inference methods share the same architecture, which is detailed in Supplement \ref{supp:vae-model}. In MNIST, we use a small-scale convolutional architecture and output Bernoulli parameters; in CIFAR10, we use a DCGAN-style architecture \citep{Radford2016UnsupervisedRL} and output Gaussian means.

\citet{pmlr-v70-hoffman17a} gives insightful training techniques: we also add an extra shearing matrix in the generative network and adapt HMC step-sizes $s$ to target an acceptance rate fixed throughout training. The best target acceptance rate for HMC in TSC and NeutraHMC is hand-tuned in [0.67, 0.95]. Number of leapfrog steps $L$ is set to $\lceil \frac{1}{s} \rceil$. The HMC initial state $\textbf{z}^{(0)}$ is sampled from the encoder, whether it is previously warmed up or not. Additionally, TSC is more computationally demanding compared with ELBO maximization because of the HMC steps. We cap $L$ to 4 to ensure similar run-time with NeutraHMC, because TSC tends to lead to smaller step-sizes. 

For TSC and NeutraHMC, we use one HMC step per data-point per epoch. For IWAE and CIS-MSC, we use 50 samples, suggested by \citet{Burda2016ImportanceWA}. 

We estimate test log-marginal likelihood $\log p(\textbf{x};\theta)$ using Annealed Importance Sampling (AIS) \citep{neal-ais, wu2017quantitative} with 10 leapfrog steps and adaptive step sizes tuned to 67\% acceptance, and 2500 annealing steps for MNIST, 7500 annealing steps for CIFAR10.

\begin{table}[h!]
\centering
\caption{Test log-marginal likelihood. Dim. refers to latent dimensions, and dim. 2 corresponds to Gaussian posterior, and dim. 64 or 128 corresponds to RealNVP posterior. -W: warm-up on the encoder previous to training. *: -2900 must be added to $\log p(\textbf{x})$ for each CIFAR10 result. \emph{TSC gives better predictive performance.}}
\subtable[Static MNIST]{
\resizebox{0.47\columnwidth}{!}{\begin{tabular}{ccc} 
 \toprule
 Dim. & Method & $\log p(\textbf{x})$ \\
 \midrule
 & ELBO VI & $-133.94$ \\
 2 & NeutraHMC-W & $-129.11$ \\
 & \textbf{TSC} & $\boldsymbol{-128.88}$ \\
 \midrule
 & ELBO VI & $-61.01$ \\ 
 & IW & $-58.84$ \\ 
 64 & NeutraHMC & $-60.44$ \\ 
 & NeutraHMC-W & $-59.62$ \\
 & CIS-MSC & $-60.4$ \\ 
 & \textbf{TSC} & $\boldsymbol{-57.97}$ \\
 \bottomrule
\end{tabular}}}
\subtable[Top: dynamic MNIST; bottom: CIFAR10]{
 \resizebox{0.47\columnwidth}{!}{\begin{tabular}{ccc} 
 \toprule
 Dim. & Method & $\log p(\textbf{x})^{\text*}$ \\
 \midrule
 & ELBO VI & $-83.7$ \\
 & IW & $-82.15$ \\
 64 & NeutraHMC-W & $-83.01$ \\
 & CIS-MSC & $-85.1$ \\
 & \textbf{TSC} & $\boldsymbol{-82.07}$ \\
 \midrule
 & ELBO VI & $-34.61$ \\
 & IW & $-33.25$ \\
 128 & NeutraHMC-W & $-33.33$ \\
 & CIS-MSC & $-33.38$ \\
 & \textbf{TSC} & $\boldsymbol{-31.23}$ \\
 \bottomrule
\end{tabular}}}
\label{tab:vae-mnist}
\end{table}

\paragraph{Results.}
TSC achieves higher log-marginal likelihood with low dimensional latent variables on Gaussian warped space and high dimensional latent variables on Real NVP warped space (Table \ref{tab:vae-mnist}). We use RealNVP instead of IAF for fast inversion $T_{\lambda}^{-1}(\textbf{z})$. Two RealNVPs are stacked to form the variational posterior, each one having two hidden layers. Every model, including baselines, uses this flow distribution, contains a single layer of latent variables, and trains for 500 epochs. 

TSC demonstrates effective synergy between transport map training and HMC sampling by training both encoders and decoders from scratch. This framework no longer requires a separate pretraining, which NeutraHMC does by warming up the encoder (which includes the normalizing flow) with ELBO maximization for 500 epochs. NeutraHMC then continues to train the warmed-up encoder during the main training phase (500 more epochs), as done in \citet{pmlr-v70-hoffman17a}. Meanwhile, in the first 10 of the 500 TSC training epochs, the encoder is not trained, a design that improves stability.


\paragraph{Ablation Studies.} 
Since we utilize both KL$(p||q)$ and a continuously-run, warped-space HMC, we wish to know whether the algorithm is as effective if one of these two components is removed. In the case of 2-dimensional latent variables, we first train a model with maximum likelihood using warped space HMC like in TSC, but it uses a pretrained encoder and no longer does KL$(p||q)$ to update the encoder. Next, we train a model that, compared to TSC, uses an ordinary HMC kernel without the space transformation. Results detailed in Supplement \ref{supp:vae} show that neither model achieves competitive performance. Therefore, not only is warped space HMC necessary for effective performance, but the dynamic KL$(p||q)$ updates of the approximate posterior and hence the transport map also play an essential role.

\section{Conclusions}

We develop Transport Score Climbing, improving VI with KL$(p||q)$ by using an HMC kernel on a simpler geometry defined by a transport map. This framework naturally leverages synergies since the transformation that warps the geometry is updated by HMC samples at each iteration, enabling more effective HMC sampling in future iterations. We illustrate the advantages of this method on two synthetic examples, survey data, and MNIST and CIFAR10 using VAE. 

\bibliographystyle{plainnat}
\bibliography{references}

\begin{thebibliography}{48}
\providecommand{\natexlab}[1]{#1}
\providecommand{\url}[1]{\texttt{#1}}
\expandafter\ifx\csname urlstyle\endcsname\relax
  \providecommand{\doi}[1]{doi: #1}\else
  \providecommand{\doi}{doi: \begingroup \urlstyle{rm}\Url}\fi

\bibitem[Abadi et~al.(2015)Abadi, Agarwal, Barham, Brevdo, Chen, Citro,
  Corrado, Davis, Dean, Devin, Ghemawat, Goodfellow, Harp, Irving, Isard, Jia,
  Jozefowicz, Kaiser, Kudlur, Levenberg, Man\'{e}, Monga, Moore, Murray, Olah,
  Schuster, Shlens, Steiner, Sutskever, Talwar, Tucker, Vanhoucke, Vasudevan,
  Vi\'{e}gas, Vinyals, Warden, Wattenberg, Wicke, Yu, and
  Zheng]{tensorflow2015-whitepaper}
Mart\'{\i}n Abadi, Ashish Agarwal, Paul Barham, Eugene Brevdo, Zhifeng Chen,
  Craig Citro, Greg~S. Corrado, Andy Davis, Jeffrey Dean, Matthieu Devin,
  Sanjay Ghemawat, Ian Goodfellow, Andrew Harp, Geoffrey Irving, Michael Isard,
  Yangqing Jia, Rafal Jozefowicz, Lukasz Kaiser, Manjunath Kudlur, Josh
  Levenberg, Dandelion Man\'{e}, Rajat Monga, Sherry Moore, Derek Murray, Chris
  Olah, Mike Schuster, Jonathon Shlens, Benoit Steiner, Ilya Sutskever, Kunal
  Talwar, Paul Tucker, Vincent Vanhoucke, Vijay Vasudevan, Fernanda Vi\'{e}gas,
  Oriol Vinyals, Pete Warden, Martin Wattenberg, Martin Wicke, Yuan Yu, and
  Xiaoqiang Zheng.
\newblock {TensorFlow}: Large-scale machine learning on heterogeneous systems,
  2015.
\newblock Software available from tensorflow.org.

\bibitem[Andrieu and Vihola(2014)]{AndrieuV:2014}
C.~Andrieu and M.~Vihola.
\newblock Markovian stochastic approximation with expanding projections.
\newblock \emph{Bernoulli}, 20\penalty0 (2), November 2014.

\bibitem[Andrieu and Moulines(2006)]{andrieu2006}
Christophe Andrieu and {\'E}ric Moulines.
\newblock {On the ergodicity properties of some adaptive MCMC algorithms}.
\newblock \emph{The Annals of Applied Probability}, 16\penalty0 (3):\penalty0
  1462 -- 1505, 2006.

\bibitem[Benveniste et~al.(1990)Benveniste, M{\'e}tivier, and
  Priouret]{benveniste2012adaptive}
Albert Benveniste, Michel M{\'e}tivier, and Pierre Priouret.
\newblock \emph{Adaptive algorithms and stochastic approximations}, volume~22.
\newblock Springer Science \& Business Media, 1990.

\bibitem[Bishop(2006)]{bishop:2006:PRML}
Christopher~M. Bishop.
\newblock \emph{Pattern Recognition and Machine Learning}.
\newblock Springer, 2006.

\bibitem[Blei et~al.(2017)Blei, Kucukelbir, and McAuliffe]{blei2017variational}
David Blei, Alp Kucukelbir, and Jon~D McAuliffe.
\newblock Variational inference: A review for statisticians.
\newblock \emph{Journal of the American Statistical Association}, 112\penalty0
  (518):\penalty0 859--877, 2017.

\bibitem[Bornschein and Bengio(2015)]{Bornschein2015}
B.~Bornschein and Yoshua Bengio.
\newblock Reweighted wake-sleep.
\newblock In \emph{International Conference on Learning Representations}, 2015.

\bibitem[Burda et~al.(2016)Burda, Grosse, and
  Salakhutdinov]{Burda2016ImportanceWA}
Yuri Burda, Roger~B. Grosse, and Ruslan Salakhutdinov.
\newblock Importance weighted autoencoders.
\newblock \emph{Computing Research Repository}, abs/1509.00519, 2016.

\bibitem[Caterini et~al.(2018)Caterini, Doucet, and
  Sejdinovic]{caterini2018hamiltonian}
Anthony~L. Caterini, A.~Doucet, and D.~Sejdinovic.
\newblock Hamiltonian variational auto-encoder.
\newblock In \emph{Neural Information Processing Systems}, 2018.

\bibitem[Dieng and Paisley(2019)]{dieng2019reweighted}
Adji~B Dieng and John Paisley.
\newblock Reweighted expectation maximization.
\newblock \emph{arXiv:1906.05850}, 2019.

\bibitem[Dillon et~al.(2017)Dillon, Langmore, Tran, Brevdo, Vasudevan, Moore,
  Patton, Alemi, Hoffman, and Saurous]{DBLP:journals/corr/abs-1711-10604}
Joshua~V. Dillon, Ian Langmore, Dustin Tran, Eugene Brevdo, Srinivas Vasudevan,
  Dave Moore, Brian Patton, Alex Alemi, Matthew~D. Hoffman, and Rif~A. Saurous.
\newblock Tensorflow distributions.
\newblock \emph{Computing Research Repository}, abs/1711.10604, 2017.

\bibitem[Dinh et~al.(2016)Dinh, Sohl{-}Dickstein, and Bengio]{dinh-realnvp}
Laurent Dinh, Jascha Sohl{-}Dickstein, and Samy Bengio.
\newblock Density estimation using real {NVP}.
\newblock \emph{Computing Research Repository}, abs/1605.08803, 2016.

\bibitem[Duane et~al.(1987)Duane, Kennedy, Pendleton, and Roweth]{duane-mcem}
Simon Duane, A.D. Kennedy, Brian~J. Pendleton, and Duncan Roweth.
\newblock Hybrid monte carlo.
\newblock \emph{Physics Letters B}, 195\penalty0 (2):\penalty0 216--222, 1987.
\newblock ISSN 0370-2693.

\bibitem[Finke and Thiery(2019)]{finke2019importanceweighted}
Axel Finke and Alexandre~H. Thiery.
\newblock On importance-weighted autoencoders.
\newblock \emph{arXiv:1907.10477}, 2019.

\bibitem[Gabrié et~al.(2021)Gabrié, Rotskoff, and
  Vanden-Eijnden]{gabrie2021adaptive}
Marylou Gabrié, Grant~M. Rotskoff, and Eric Vanden-Eijnden.
\newblock Adaptive {M}onte {C}arlo augmented with normalizing flows.
\newblock \emph{arXiv:2105.12603}, 2021.

\bibitem[Gu and Kong(1998)]{Gu7270}
Ming~Gao Gu and Fan~Hui Kong.
\newblock A stochastic approximation algorithm with {M}arkov chain
  {M}onte-{C}arlo method for incomplete data estimation problems.
\newblock \emph{Proceedings of the National Academy of Sciences}, 95\penalty0
  (13):\penalty0 7270--7274, 1998.

\bibitem[Gu et~al.(2015)Gu, Ghahramani, and Turner]{gu2015neural}
Shixiang~(Shane) Gu, Zoubin Ghahramani, and Richard~E Turner.
\newblock Neural adaptive sequential {M}onte {C}arlo.
\newblock In \emph{Neural Information Processing Systems}, pages 2629--2637.
  Curran Associates, Inc., 2015.

\bibitem[Haario et~al.(1999)Haario, Saksman, and Tamminen]{haario1999}
Heikki Haario, Eero Saksman, and Johanna Tamminen.
\newblock Adaptive proposal distribution for random walk metropolis algorithm.
\newblock \emph{Computational Statistics}, 14\penalty0 (3):\penalty0 375--395,
  1999.

\bibitem[Hoffman et~al.(2013)Hoffman, Blei, Wang, and Paisley]{Hoffman2013}
M.~D. Hoffman, D.~Blei, C.~Wang, and J.~Paisley.
\newblock Stochastic variational inference.
\newblock \emph{Journal of Machine Learning Research}, 14:\penalty0 1303--1347,
  May 2013.

\bibitem[Hoffman(2017)]{pmlr-v70-hoffman17a}
Matthew~D. Hoffman.
\newblock Learning deep latent {G}aussian models with {M}arkov chain {M}onte
  {C}arlo.
\newblock In Doina Precup and Yee~Whye Teh, editors, \emph{Proceedings of the
  34th International Conference on Machine Learning}, volume~70 of
  \emph{Proceedings of Machine Learning Research}, pages 1510--1519. PMLR,
  06--11 Aug 2017.

\bibitem[Hoffman et~al.(2019)Hoffman, Sountsov, Dillon, Langmore, Tran, and
  Vasudevan]{hoffman2019neutralizing}
Matthew~D. Hoffman, Pavel Sountsov, Joshua~V. Dillon, Ian Langmore, Dustin
  Tran, and Srinivas Vasudevan.
\newblock Neutra-lizing bad geometry in hamiltonian monte carlo using neural
  transport.
\newblock \emph{arXiv: Computation}, 2019.

\bibitem[Jerfel et~al.(2021)Jerfel, Wang, Wong-Fannjiang, Heller, Ma, and
  Jordan]{jerfel2021variational}
Ghassen Jerfel, Serena Wang, Clara Wong-Fannjiang, Katherine~A. Heller, Yian
  Ma, and Michael~I. Jordan.
\newblock Variational refinement for importance sampling using the forward
  kullback-leibler divergence.
\newblock In \emph{Proceedings of the Thirty-Seventh Conference on Uncertainty
  in Artificial Intelligence}, volume 161 of \emph{Proceedings of Machine
  Learning Research}, pages 1819--1829. PMLR, 27--30 Jul 2021.

\bibitem[Jordan et~al.(1999)Jordan, Ghahramani, Jaakkola, and Saul]{Jordan1999}
M.~I. Jordan, Z.~Ghahramani, T.~S. Jaakkola, and L.~K. Saul.
\newblock An introduction to variational methods for graphical models.
\newblock \emph{Machine Learning}, 37\penalty0 (2):\penalty0 183--233, November
  1999.

\bibitem[Kim et~al.(2022)Kim, Oh, Gardner, Dieng, and Kim]{kim2022markov}
Kyurae Kim, Jisu Oh, Jacob~R. Gardner, Adji~Bousso Dieng, and Hongseok Kim.
\newblock Markov chain score ascent: A unifying framework of variational
  inference with {M}arkovian gradients.
\newblock \emph{arXiv:2206.06295}, 2022.

\bibitem[Kingma and Ba(2015)]{Kingma2015AdamAM}
Diederik~P. Kingma and Jimmy Ba.
\newblock Adam: A method for stochastic optimization.
\newblock \emph{Computing Research Repository}, abs/1412.6980, 2015.

\bibitem[Kingma and Welling(2014)]{Kingma2014AutoEncodingVB}
Diederik~P. Kingma and M.~Welling.
\newblock Auto-encoding variational {B}ayes.
\newblock \emph{Computing Research Repository}, abs/1312.6114, 2014.

\bibitem[Kingma et~al.(2016)Kingma, Salimans, and Welling]{kingma-iaf}
Diederik~P. Kingma, Tim Salimans, and Max Welling.
\newblock Improving variational inference with inverse autoregressive flow.
\newblock \emph{Computing Research Repository}, abs/1606.04934, 2016.

\bibitem[Kuhn and Lavielle(2004)]{KuhnL:2004}
E.~Kuhn and M.~Lavielle.
\newblock Coupling a stochastic approximation version of {EM} with an {MCMC}
  procedure.
\newblock \emph{European Series in Applied and Industrial Mathematics:
  Probability and Statistics}, 8:\penalty0 115--131, 2004.

\bibitem[Lopez-Martin et~al.(2021)Lopez-Martin, Phillips, and
  Gelman]{lopez-martin_phillips_gelman_2021}
Juan Lopez-Martin, Justin~H. Phillips, and Andrew Gelman.
\newblock Multilevel regression and poststratification case studies, Feb 2021.
\newblock URL \url{https://bookdown.org/jl5522/MRP-case-studies/}.

\bibitem[Mangoubi and Smith(2017)]{mangoubi2017rapid}
Oren Mangoubi and Aaron Smith.
\newblock Rapid mixing of hamiltonian monte carlo on strongly log-concave
  distributions.
\newblock \emph{arXiv: Probability}, 2017.

\bibitem[Marzouk et~al.(2016)Marzouk, Moselhy, Parno, and
  Spantini]{Marzouk_2016}
Youssef Marzouk, Tarek Moselhy, Matthew Parno, and Alessio Spantini.
\newblock Sampling via measure transport: An introduction.
\newblock \emph{Handbook of Uncertainty Quantification}, page 1–41, 2016.

\bibitem[Minka(2005)]{minka2005divergence}
Tom Minka.
\newblock Divergence measures and message passing.
\newblock Technical report, Technical report, Microsoft Research, 2005.

\bibitem[Naesseth et~al.(2019)Naesseth, Lindsten, and
  Schön]{naesseth2019elements}
Christian~A. Naesseth, Fredrik Lindsten, and Thomas~B. Schön.
\newblock Elements of sequential {M}onte {C}arlo.
\newblock \emph{Foundations and Trends® in Machine Learning}, 12\penalty0
  (3):\penalty0 307--392, 2019.

\bibitem[Naesseth et~al.(2020)Naesseth, Lindsten, and Blei]{Naesseth2020}
Christian~A. Naesseth, Fredrik Lindsten, and David Blei.
\newblock Markovian score climbing: Variational inference with {KL}(p$\|$q).
\newblock In \emph{Neural Information Processing Systems}, 2020.

\bibitem[Neal(2001)]{neal-ais}
Radford~M. Neal.
\newblock Annealed importance sampling.
\newblock \emph{Statistics and Computing}, 11\penalty0 (2):\penalty0 125--139,
  2001.

\bibitem[Neal(2003)]{neal-2003}
Radford~M. Neal.
\newblock {Slice sampling}.
\newblock \emph{The Annals of Statistics}, 31\penalty0 (3):\penalty0 705 --
  767, 2003.

\bibitem[Neal(2011)]{neal-hmc}
Radford~M. Neal.
\newblock \emph{Handbook of Markov Chain Monte Carlo}.
\newblock Chapman and Hall/CRC, May 2011.
\newblock ISBN 9780429138508.

\bibitem[Ou and Song(2020)]{ou2020joint}
Zhijian Ou and Yunfu Song.
\newblock Joint stochastic approximation and its application to learning
  discrete latent variable models.
\newblock In \emph{Conference on Uncertainty in Artificial Intelligence}, 2020.

\bibitem[Radford et~al.(2016)Radford, Metz, and
  Chintala]{Radford2016UnsupervisedRL}
Alec Radford, Luke Metz, and Soumith Chintala.
\newblock Unsupervised representation learning with deep convolutional
  generative adversarial networks.
\newblock \emph{Computing Research Repository}, abs/1511.06434, 2016.

\bibitem[Rezende and Mohamed(2015)]{rezende2015flow}
Danilo~Jimenez Rezende and Shakir Mohamed.
\newblock Variational inference with normalizing flows.
\newblock In \emph{International Conference on Machine Learning}, 2015.

\bibitem[Rezende et~al.(2014)Rezende, Mohamed, and
  Wierstra]{Rezende2014StochasticBA}
Danilo~Jimenez Rezende, Shakir Mohamed, and Daan Wierstra.
\newblock Stochastic backpropagation and approximate inference in deep
  generative models.
\newblock In \emph{International Conference on Machine Learning}, 2014.

\bibitem[Robert and Casella(2004)]{robert2013monte}
Christian Robert and George Casella.
\newblock \emph{Monte {C}arlo Statistical Methods}.
\newblock Springer Science \& Business Media, 2004.

\bibitem[Ruiz et~al.(2021)Ruiz, Titsias, taylan. cemgil, and
  Doucet]{ruiz2021unbiased}
Francisco J.~R. Ruiz, Michalis~K. Titsias, taylan. cemgil, and A.~Doucet.
\newblock Unbiased gradient estimation for variational auto-encoders using
  coupled markov chains.
\newblock In \emph{Conference on Uncertainty in Artificial Intelligence}, 2021.

\bibitem[Salimans et~al.(2015)Salimans, Kingma, and
  Welling]{salimans2015markov}
Tim Salimans, Diederik~P. Kingma, and Max Welling.
\newblock Markov chain monte carlo and variational inference: Bridging the gap.
\newblock In \emph{International Conference on Machine Learning}, 2015.

\bibitem[Schaffner et~al.(2019)Schaffner, Ansolabehere, and
  Luks]{DVN/ZSBZ7K_2019}
Brian Schaffner, Stephen Ansolabehere, and Sam Luks.
\newblock {CCES Common Content, 2018}.
\newblock 2019.

\bibitem[Tabak and Turner(2013)]{tabak2013family}
Esteban~G Tabak and Cristina~V Turner.
\newblock A family of nonparametric density estimation algorithms.
\newblock \emph{Communications on Pure and Applied Mathematics}, 66\penalty0
  (2):\penalty0 145--164, 2013.

\bibitem[Wu et~al.(2017)Wu, Burda, Salakhutdinov, and
  Grosse]{wu2017quantitative}
Yuhuai Wu, Yuri Burda, Ruslan Salakhutdinov, and Roger~B. Grosse.
\newblock On the quantitative analysis of decoder-based generative models.
\newblock \emph{ArXiv}, abs/1611.04273, 2017.

\bibitem[Yao et~al.(2018)Yao, Vehtari, Simpson, and Gelman]{Yao2018YesBD}
Yuling Yao, Aki Vehtari, Daniel~P. Simpson, and Andrew Gelman.
\newblock Yes, but did it work?: Evaluating variational inference.
\newblock In \emph{International Conference on Machine Learning}, 2018.

\end{thebibliography}

\newpage
\appendix
\onecolumn
\section*{Appendix}

\section{Convergence of \textsc{TSC}}\label{sec:convergence}
The \textsc{TSC} parameter estimates, $\lambda$ and $\theta$, may converge to a local optima of the forward KL and marginal likelihood, respectively. We formalize this result and detail the conditions in Proposition~\ref{prop:convergence} for $\xi=(\theta,\lambda)$. The proposition is an application of \citep[Theorem 1]{Gu7270} which relies on \citep[Theorem 3.17, page 304]{benveniste2012adaptive}, and an adaptation of the result by \citet{Naesseth2020} that focuses only on $\lambda$. 
\begin{prop}
Assume A\ref{cond:1}-\ref{cond:6}, defined in the Supplement. If $\xi_k = (\theta_k,\lambda_k)$ for $k \geq 1$, defined in Algorithm~\ref{alg:hsc}, is a bounded sequence that almost surely visits a compact subset of the domain of attraction of $\xi^\star=(\theta^\star,\lambda^\star)$ infinitely often, then
$$\xi_k \to \xi^\star, \quad \text{almost surely}.$$ 
\label{prop:convergence}
\end{prop}

The proposition is an adaptation of \citet[Theorem 1]{Gu7270} based on \citet[Theorem 3.17, page 304]{benveniste2012adaptive} and a minor extension of \citet[Proposition 1]{Naesseth2020}. Let $\xi^\star = (\theta^\star,\lambda^\star)$, where $\theta^\star$ is a maximizer of the log-marginal likelihood and $\lambda^\star$ is a minimizer of the forward KL divergence. Consider the ordinary differential equation (ODE), for $\xi(t) = (\theta(t),\lambda(t))$, defined by
\begin{align}
	\frac{\mathrm{d}}{\mathrm{d}t} \xi(t) &= 
	\left(
	\begin{array}{c}
	\mathbb{E}_{p({\bf z|x;\theta})} [ \nabla_{\theta}\log p({\bf z, x};\theta(t)) ] \\
	\mathbb{E}_{p({\bf z|x;\theta})} [ \nabla_{\lambda} \log q(\textbf{z};\lambda(t)) ]
	\end{array}
	\right), ~ \xi(0) = \xi_0,
	\label{eq:ode}
\end{align}
and its solution $\xi(t)$ for $t \geq 0$. If $\xi(t)=\widehat{\xi}$ is an unique solution to eq.~\ref{eq:ode} for $\xi_0=\widehat{\xi}$, we call $\widehat{\xi}$ a stability point. The optima $\xi^\star$ is a stability point for eq.~\ref{eq:ode}. We call the set $\Xi$ a domain of attraction of $\widehat{\xi}$, if the solution of eq.~\ref{eq:ode} for $\xi_0\in\Xi$ remains in $\Xi$ and converges to $\widehat{\xi}$. Suppose that $\Xi$ is an open set in $\mathbb{R}^{d_\xi}$ and that $\xi_k \in \mathbb{R}^{d_\xi}$. Further, suppose ${\bf z_k} \in \mathbb{R}^{d_z}$ and that $\mathrm{Z}$ is an open set in $\mathbb{R}^{d_z}$. Denote the Hamiltonian Markov kernel used in TSC by $H_\xi({\bf z},\mathrm{d}{\bf z'})$,and repeated application of this kernel $H_\xi^k({\bf z},\mathrm{d}{\bf z'})=\int\cdots\int H_\xi({\bf z},\mathrm{d}{\bf z_1})\cdots H_\xi({\bf z}_{k-1},\mathrm{d}{\bf z'})$. The length of the vector ${\bf z}$ is denoted by $|{\bf z}|$. Let $Q$ be any compact subset of $\Xi$, and $q>1$ a sufficiently large (real) number so that the following assumptions holds. Like \citet{Gu7270} we assume:
\begin{cond}
The step size sequence satisfies $\sum_{k=1}^\infty \alpha_k = \infty$ and $\sum_{k=1}^\infty \alpha_k^2 < \infty$.
\label{cond:1}
\end{cond}
\begin{cond}[Integrability]
There exists a constant $c_1$ such that for any $\xi \in \Xi$, $\mathbf{z} \in \mathrm{Z}$ and $k \geq 1$  
\begin{align*}
    \int \left(|{\bf z}|^q+1\right)H_\xi^k({\bf z},\mathrm{d}{\bf z'}) \leq c_1 \left(|{\bf z}|^q+1\right)
\end{align*}
\label{cond:2}
\end{cond}
\begin{cond}[Markov Chain Convergence]
Let $p({\bf z|x;\theta})$ be the unique invariant distribution for $H_\xi$. For each $\xi \in \Xi$
\begin{align*}
    \lim_{k\to\infty} \sup_{{\bf z}\in\mathrm{Z}} \frac{\int \left(|{\bf z'}|^q+1\right)|H_\xi^k({\bf z},\mathrm{d}{\bf z'})-p(\mathrm{d}{\bf z'|x;\theta})|}{|{\bf z}|^q+1} = 0
\end{align*}
\label{cond:3}
\end{cond}
\begin{cond}[Continuity in $\xi$]
There exists a constant $c_2$ such that for all $\xi,\xi' \in Q$, 
\begin{align*}
    \int \left(|{\bf z'}|^q+1\right)|H_\xi({\bf z},\mathrm{d}{\bf z'})-H_{\xi'}({\bf z},\mathrm{d}{\bf z'})| \leq c_2 |\xi-\xi'| \left(|{\bf z}|^q+1\right)
\end{align*}
\label{cond:4}
\end{cond}
\begin{cond}[Continuity in ${\bf z}$]
There exists a constant $c_3$ such that for all ${\bf z}_1,{\bf z}_2 \in \mathrm{Z}$  
\begin{align*}
    \sup_{\xi \in \Xi} \left| \int \left(|{\bf z'}|^q+1\right) \left(H_\xi({\bf z}_1,\mathrm{d}{\bf z'})-H_{\xi'}({\bf z}_2,\mathrm{d}{\bf z'})\right) \right| \leq c_3 |{\bf z}_1-{\bf z}_2|\left(|{\bf z}_1|^q+|{\bf z}_2|^q +1\right)
\end{align*}
\label{cond:5}
\end{cond}
\begin{cond}[Conditions on Gradients]
For any compact subset $Q \subset \Xi$, there exists (positive) constants $p$, $k_1$, $k_2$, $k_3$, $\nu > \frac{1}{2}$ such that for all $\xi, \xi'\in \Xi$ and ${\bf z},{\bf z}_1,{\bf z}_2 \in \mathrm{Z}$
\begin{align*}
    |\nabla_\xi\left(\log p({\bf z, x};\theta)+\log q({\bf z};\lambda)\right)| &\leq k_1 \left(|{\bf z}|^{p+1}+1\right) \\
    |\nabla_\xi\left(\log p({\bf z}_1, {\bf x};\theta)+\log q({\bf z}_1;\lambda)\right)-\nabla_\xi\left(\log p({\bf z}_2, {\bf  x};\theta)+\log q({\bf z_2};\lambda)\right)| &\leq k_2 |{\bf z}_1-{\bf z}_2| \left(|{\bf z}_1|^{p}+|{\bf z}_2|^{p}+1\right) \\
    |\nabla_\xi\left(\log p({\bf z, x};\theta)+\log q({\bf z};\lambda)\right)-\nabla_\xi\left(\log p({\bf z, x};\theta')+\log q({\bf z};\lambda')\right)| &\leq k_3 |\xi-\xi'|^\nu \left(|{\bf z}|^{p+1}+1\right)
\end{align*}
\label{cond:6}
\end{cond}

The results follows from \citet[Theorem 1]{Gu7270}, under assumptions A\ref{cond:1}-\ref{cond:6}, by identifying:
\begin{align*}
    \theta &= \xi \\
    x &= {\bf z} \\
    \Pi_\theta &= H_\xi \\
    H(\theta,x) &= \nabla_\xi\left(\log p({\bf z, x};\theta)+\log q({\bf z};\lambda)\right)
\end{align*}
and $\Gamma_k=I$, $I(\theta,x)=0$ where left is their notation and right is our notation.

\begin{figure*}[h!]
    \centering
    \subfigure[TSC, MRP MCMC, and 60000 sample unadjusted estimate.]{
    \centering
    \includegraphics[width=0.85\textwidth]{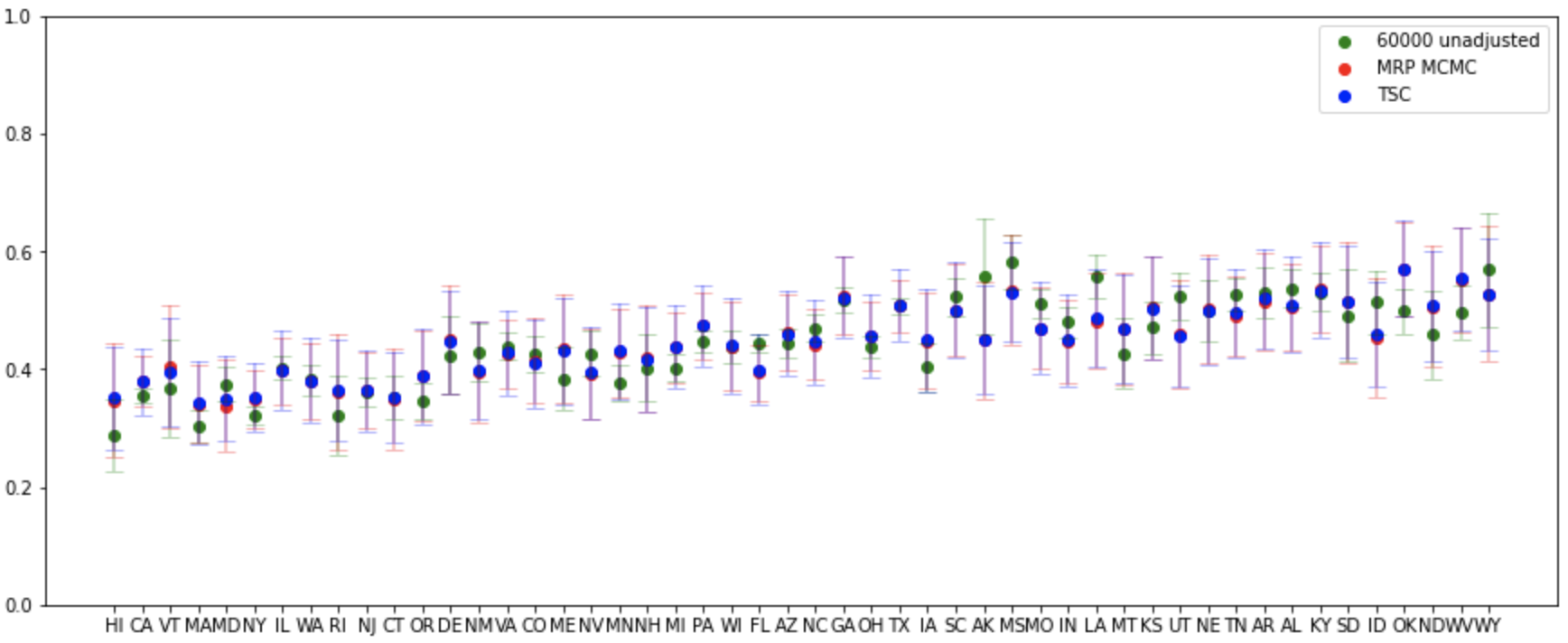}}
    \subfigure[TSC, MSC, and 60000 sample unadjusted estimate.]{
    \centering
    \includegraphics[width=0.85\textwidth]{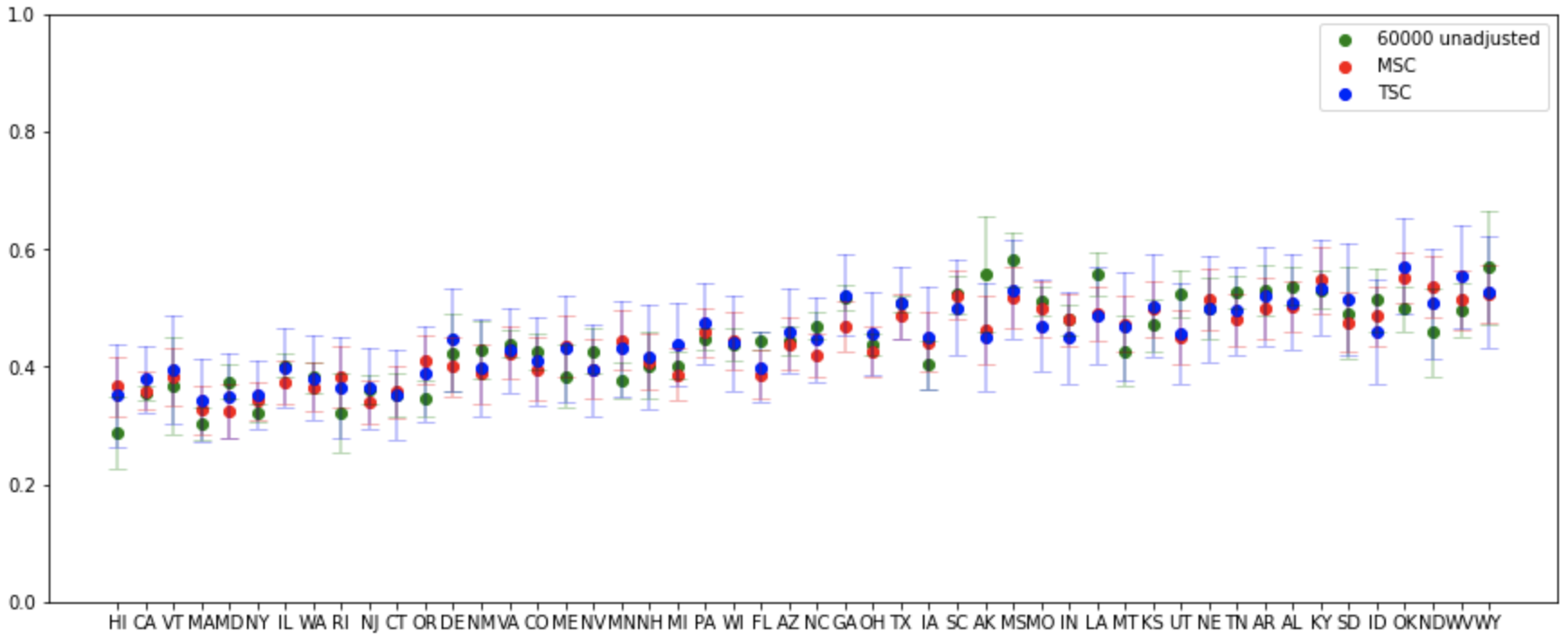}}
    \caption{Estimates by states, where states are ordered by Republican vote in the 2016 election. 95\% confidence intervals are plotted. The closer to green the better.}
    \label{fig:survey-by-state-more}
\end{figure*}

\section{Experiments}

\subsection{Survey Data}
\paragraph{Model.} \label{supp:survey-model}
Following \citep{lopez-martin_phillips_gelman_2021}, we model binary variable \textbf{x} taking values 0 or 1 with a multilevel regression model. \textbf{x} indicate individual responses, and each individual comes with given features: state, age, ethnicity, education, and gender. For each data-point $x_i$, the model is defined as,
\begin{align*}
    p(x_i=1) &= \text{logit}^{-1}(z^{\text{state}}_{s[i]} + z^{\text{age}}_{a[i]} + z^{\text{eth}}_{r[i]} + z^{\text{edu}}_{e[i]} + z^{\text{gen.eth}}_{g[i],r[i]} + z^{\text{edu.age}}_{e[i],a[i]} + z^{\text{edu.eth}}_{e[i],r[i]} + \beta^{\text{gen}} \cdot \text{Gen}_i). \\
    z^{\text{state}}_{s} &\sim \mathcal{N}(\gamma^0 + \gamma^{\text{south}}\cdot \mathbbm{1}(\text{s in south}) + \gamma^{\text{northcentral}}\cdot \mathbbm{1}(\text{s in northcentral}) + \gamma^{\text{west}}\cdot \mathbbm{1}(\text{s in west}), \sigma^{\text{state}}), \text{for $s$ = 1, ..., 50}, \\
    z^{\text{age}}_{a} &\sim \mathcal{N}(0, \sigma^{\text{age}}), \text{for $a$ = 1, ..., 6}, \\
    z^{\text{eth}}_{r} &\sim \mathcal{N}(0, \sigma^{\text{eth}}), \text{for $r$ = 1, ..., 4}, \\
    z^{\text{edu}}_{e} &\sim \mathcal{N}(0, \sigma^{\text{edu}}), \text{for $e$ = 1, ..., 5}, \\
    z^{\text{gen.eth}}_{g,r} &\sim \mathcal{N}(0, \sigma^{\text{gen.eth}}), \text{for $g$ = 1, 2 and $r$ = 1, ..., 4}, \\
    z^{\text{edu.age}}_{e,a} &\sim \mathcal{N}(0, \sigma^{\text{edu.age}}), \text{for $e$ = 1, ..., 5 and $a$ = 1, ..., 6}, \\
    z^{\text{edu.eth}}_{e,r} &\sim \mathcal{N}(0, \sigma^{\text{edu.eth}}), \text{for $e$ = 1, ..., 5 and $r$ = 1, ..., 4}.
\end{align*}
Each $z_{\text{*}}^{\text{*}}$ is a latent variable. For example, $z_{:}^{\text{state}}$ is a length-50 latent variable that indicates the effect of state on the binary response. As another example, $z_{:,:}^{\text{edu.age}}$ indicates the interaction effect of education and age, and is length-30 because there are 5 education levels and 6 age levels. In total, the model has a length-123 latent variable \textbf{z}. We model the rest, namely $\gamma^{\text{*}}$, $\sigma^{\text{*}}$, and $\beta^{\text{gen}}$, as model parameters where we find fixed estimates.

\paragraph{Results.} Reliable approximations on the survey data are shown by ability to generalize from small, 5,000 sample to the full 60,000 sample, and closeness to gold-standard MRP MCMC results. We visualize TSC, MSC, and MRP MCMC estimates by state. Figure \ref{fig:survey-by-state-more}a shows that the mean of TSC estimates are barely discernible from the mean of results given by the gold-standard MRP MCMC \citep{lopez-martin_phillips_gelman_2021}. Figure \ref{fig:survey-by-state-more}b shows that MSC is also robust against noise from the small 5,000 sample, but it slightly differs in results from TSC.

\subsection{Variational Autoencoder} \label{supp:vae}

\paragraph{Architecture.} \label{supp:vae-model} In MNIST, both statically and dynamically binarized, the encoder uses two convolutional layers with number of filters 32 and 64, followed by a dense layer that outputs Gaussian mean and log-variances (so its hidden-size is two times latent variable dimension). The decoder begins with a dense layer with hidden-size $7 \cdot 7 \cdot 32$, followed by three transpose convolutional layers with number of filters 32, 64, and 1, and it outputs a Bernoulli parameter for each pixel. All layers use kernel size 3, stride size 2, same padding, and ReLU activations, except for the last transpose convolutional layer that uses stride size 1.

A DCGAN-style architecture is used for CIFAR10, featuring no dense layers, batch normalization, and leaky ReLU. The encoder uses four convolutional layers with number of filters 64, 128, 256, and latent dimension times 2. The last layer has no activation function and is flattened to give Gaussian mean and log-variances. The decoder uses four transpose convolutional layers with number of filters 256, 128, 64, 3. The last layer uses tanh activation and outputs Gaussian mean. Batch normalization and leaky ReLU (0.2) are applied after each layer except for the last layer in encoder and decoder. All layers use kernel size are 4, stride size 2, and same padding, except that the last layer in encoder and first layer in decoder use stride size 1 and valid padding.


\paragraph{Ablation Studies.} We do two VAE ablation studies under the case of two dimensional latent variables.

Study I: no approximate inference with KL$(p||q)$; only maximum likelihood on $p(\textbf{x};\theta)$. First, we wonder whether warped space HMC itself along with a pretrained transport map achieves competitive performance. That is, the encoder is no longer trained, and the overall training is essentially Monte Carlo EM \citep{duane-mcem,Kingma2014AutoEncodingVB}. It achieves $-156.1$ log-marginal likelihood after the same number of epochs of training, lower than all baselines.

Study II: run HMC on original space instead of warped space. We also test whether running HMC on the original space together with approximate posterior training via KL$(p||q)$ achieves competitive performance. The estimated log-marginal likelihood is $-133.9$, lower than both TSC and NeutraHMC.

\end{document}